\documentclass[sigconf]{acmart}

\usepackage{multicol,graphicx,mathrsfs,amsmath,pifont,amscd,latexsym,color,fancyhdr,CJK,url,longtable, pifont,epstopdf,setspace,multirow,colortbl,tabularx,threeparttable, booktabs, verbatim, bbm,listings, balance,color,indentfirst,xspace,hyperref,anyfontsize,mathtools,subcaption}

\usepackage[linesnumbered,boxed,algosection]{algorithm2e}
\SetKwComment{tcp}{$\triangleright$ }{}

\SetCommentSty{mycommfont}

\usepackage{array}
\newcolumntype{L}[1]{>{\raggedright\let\newline\\\arraybackslash\hspace{0pt}}m{#1}}
\newcolumntype{C}[1]{>{\centering\let\newline\\\arraybackslash\hspace{0pt}}m{#1}}
\newcolumntype{R}[1]{>{\raggedleft\let\newline\\\arraybackslash\hspace{0pt}}m{#1}}
\newcommand{\argmin}[1]{\underset{#1}{\operatorname{arg}\,\operatorname{min}}\;}

\renewcommand\footnotetextcopyrightpermission[1]{} % removes footnote with conference information in first column
\setcopyright{acmcopyright}
\usepackage[mathscr]{euscript}
\let\euscr\mathscr \let\mathscr\relax% just so we can load this and rsfs

\usepackage{amsmath}
\DeclareMathOperator*{\argmax}{arg\,max}

\newcommand{\sys}{\texttt{DecNAS}}
\newcommand{\multipliers}{Multipliers}

\begin{document}
\title{Federated Neural Architecture Search}

 \author{Jinliang Yuan$^1$, Mengwei Xu$^1$, Yuxin Zhao$^2$, Kaigui Bian$^2$, Gang Huang$^2$, Xuanzhe Liu$^2$, Shangguang Wang$^1$}
 \affiliation {
 	\institution{$^1$State Key Laboratory of Networking and Switching Technology (BUPT), Beijing, China}
 	\country{}
 }
 \affiliation {
 	\institution{$^2$Key Lab of High Confidence Software Technologies (Peking University), Beijing, China}
 	\country{}
 }

\begin{abstract}
% !TeX root = main.tex

To preserve user privacy while enabling mobile intelligence,  techniques have been proposed to train deep neural networks on  decentralized data.
However, decentralized training makes the design of neural architecture quite difficult as it already was.
Such difficulty is further amplified when designing and deploying different neural architectures for heterogeneous mobile platforms.
In this work, we propose an automatic neural architecture search into the decentralized training, as a new DNN training paradigm called Federated Neural Architecture Search, namely federated NAS.
To deal with the primary challenge of limited on-client computational and communication resources,
we present \sys{}, a highly optimized framework for efficient federated NAS.
\sys{} fully exploits the key opportunity of insufficient model candidate re-training during the architecture search process,
and incorporates three key optimizations: parallel candidates training on partial clients, early dropping candidates with inferior performance, and dynamic round numbers.
Tested on large-scale datasets and typical CNN architectures,
\sys{} achieves comparable model accuracy as state-of-the-art NAS algorithm that trains models with centralized data,
and also reduces the client cost by up to 200$\times$ or more compared to a straightforward design of federated NAS.
\end{abstract}

\maketitle

% \keywords{Deep Learning, Mobile Vision, Cache}

%\noindent \textbf{TODO:}

%\note{AutoML (more generic) vs. NAS (our target)}

%\note{Framework vs. Algorithm vs. Protocol}

%\note{auto tuning the round number. based on the delta compared to the previous global model?}
\section{Introduction}

Attentions have been recently put onto the privacy concerns in the machine learning pipeline,
especially the deep neural networks (DNNs), which are increasingly adopted on mobile devices~\cite{xu2019first,kdd/WangZBZCY18} and often require a large amount of sensitive data (e.g., images and input corpus) from mobile users to train.
As one of the many typical examples, the recent release of  General Data Protection Regulation (GDPR)~\cite{gdpr} by European Union strictly regulates whether and how companies can access the personal data owned by their users.  In parallel, a lot of efforts have been made in the research community to design novel paradigm of machine learning and large-scale data mining that preserves the privacy of end users. One promising direction is the emerging federated learning~\cite{mcmahan2016communication}, which aims to train DNN models in a decentralized way, collaboratively from the contribution of many client devices, without gathering the private data from individual devices to the cloud.

\begin{figure}[t]
	\centering
	\includegraphics[width=0.47\textwidth]{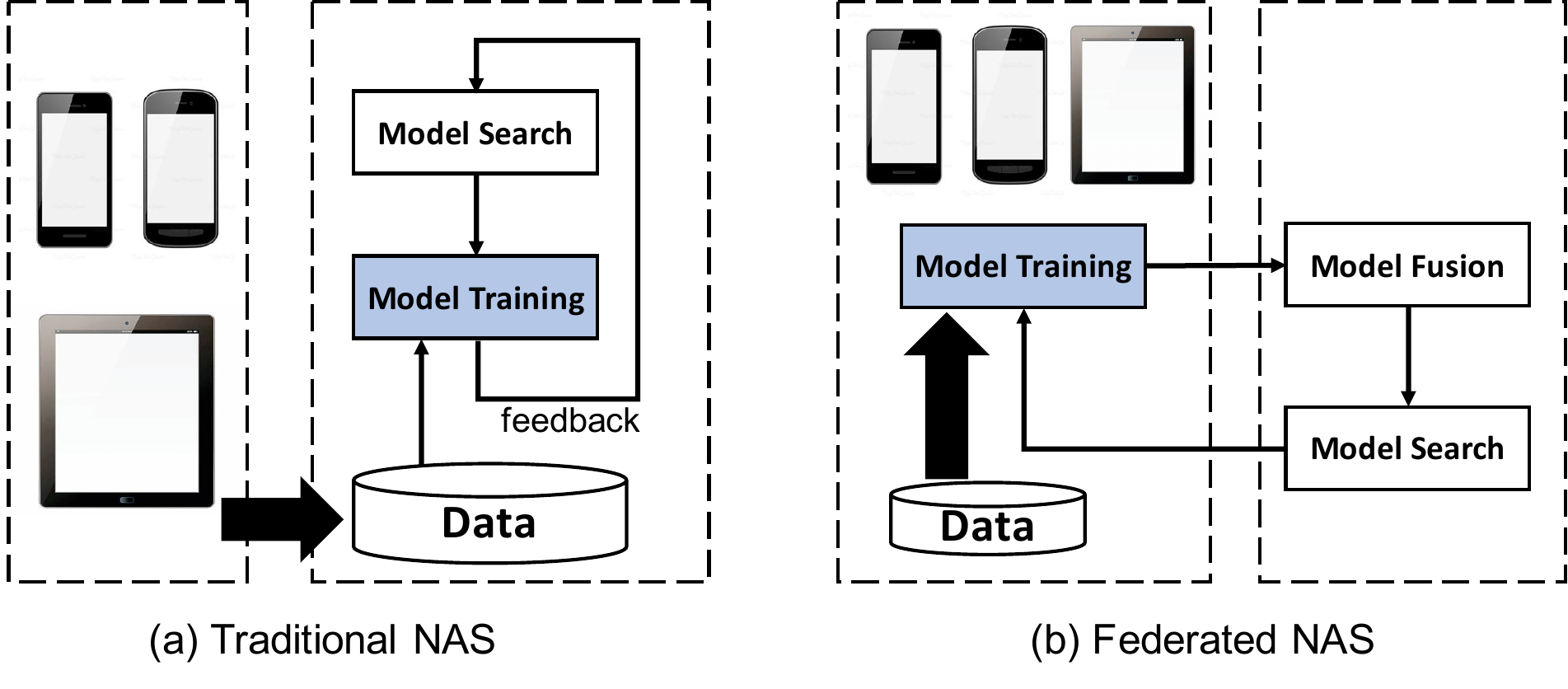}
	%\vspace*{-.6cm}
	\caption{A high-level comparison between traditional centralized NAS and federated NAS.}
	\label{fig:arch}
	\vspace{-0.3cm}
\end{figure}

Training neural network models on decentralized data and devices addresses the privacy issue to some extent at the expense of efficiency.
Once a neural network architecture is determined, there are newly developed methods to speed up the decentralized training process~\cite{konevcny2016federated,mcmahan2016communication}.
However, in most tasks when the network architecture is not determined a priori, it remains very difficult to search for the optimal network architecture(s) and train them efficiently in a decentralized setup.
Indeed, it is well known that designing an efficient architecture is a labor-intensive process that may require a vast number of iterations of training attempts, which could become uninhibitedly  time-consuming given decentralized training data.
Making it worse, the hardware platforms on mobile devices are highly heterogeneous~\cite{AI-benchmark}, thus different network architectures are required to manage diverse resource budgets on the hardware. This becomes a bottleneck of practically deploying federated learning, given the increasingly important role of neural architecture search (or NAS) in launching deep learning in reality.  

To address this major challenge, this paper proposes a new paradigm for DNN training to enable automatic neural architecture search (NAS) on decentralized data, called \textit{federated NAS}.
The major goal is to address both automation and privacy issues while training DNNs with heterogeneous mobile devices.
As shown in Figure~\ref{fig:arch}, the basic guiding idea of federated NAS is to decouple the two primary logic steps of NAS process, i.e., model search and model training, and separately distribute them on cloud and clients.
Specifically, every single client uses only its local dataset to train and test a model, while the cloud coordinates all the clients and determines the searching direction without requiring the raw data.

Given the preceding conceptual principles, enabling Neural architecture search in a federated setting is fundamentally challenging due to  limited on-client hardware resources, i.e., computation and communication.
 NAS is known to be computation-intensive (e.g., thousands of GPU-hrs~\cite{rl-nas}), given the large number of model candidates to be explored. Meanwhile, the communication cost between cloud and clients also scales up with the increased  number of model candidates. 
 It is also worth mentioning that in the federated NAS paradigm, the data distribution among clients are often non-iid and highly-skewed~\cite{mcmahan2016communication}, which can probably mislead the NAS algorithm to select non-optimal DNN candidate.

%\textbf{\sys{} framework}
We present a framework for federated NAS, named \textit{\sys}.
\sys{} starts from an expensive pre-trained model, and iteratively adapts the model to a more compact one until it meets a user-specified resource budget.
For each iteration, \sys{} generates a list of pruned model candidates, then re-trains (tunes) and tests them collaboratively across the cloud and clients.
The most accurate one will be selected before moving to the next iteration.
When terminated, \sys{} outputs a sequence of simplified DNN architectures that form the efficient frontier that strikes a balance on the trade-off of model accuracy and resource consumption. 

%\sys{} is able to reduce the on-client cost by by addressing the problem of \textit{insufficient candidate tuning during search process}.
%While previous work~\cite{netadapt,liu2018progressive,cai2018path,tan2019mnasnet} realized that the candidate tuning can be short-term without getting converged, they did not investigate how long the tuning process should be sufficient. 

By learning and retrofitting the idea of using proxy task as insufficient candidate re-training from the previous work~\cite{netadapt,liu2018progressive,cai2018path,tan2019mnasnet}, \sys{} provides several insightful mechanisms, i.e., the parallel tuning of each DNN candidate (across clients), dynamic (across time), and heterogeneous (across models), to make federated NAS practical. 
\textbf{(1)} By parallel tuning, \sys{} works on different model candidates simultaneously and recognizes the available clients into many \textit{groups}. All clients in a group  collaboratively train and test a DNN candidate with their results (accuracy, gradients, etc) uploaded and properly fused on the cloud. Different groups work on different DNN candidates in parallel to increase the scalability by involving more available clients.
%The reason of one DNN candidate only involves a subset of clients is because the state-of-the-art gradients fusion algorithm, \texttt{FedAvg}, cannot efficiently utilize more than hundreds of devices in parallel~\cite{bonawitz2019towards}.
To ensure the generality of each DNN candidate, \sys{} incorporates a principled client partition algorithm with regard to each client' data distribution and data size.
\textbf{(2)} By dynamic training, \sys{} increasingly trains each candidate with more rounds as iterations go on, instead of using a fixed and large round number as prior NAS works~\cite{netadapt}.
This is based on the observation that as the model being simplified to smaller, each DNN candidate requires more re-training to regain the accuracy so that \sys{} can adapt the model at the right direction.
\textbf{(3)} By model heterogeneity, \sys{} early drops the non-optimal candidates during the re-training stages (e.g., 2 rounds), but only the optimal one is trained for the required round number (e.g., 10).
This is based on the observation that the optimal DNN candidate often quickly outperforms others even far before the re-training is done.

We comprehensively evaluated the performance of \sys{} on two datasets, ImageNet (iid) and Celeba (non-iid), as well as two CNN architectures, i.e., MobileNet and simplified AlexNet. The results show that
\sys{} achieves similar model accuracy as state-of-the-art NAS algorithm that trains models on centralized data, and the three novel optimizations above can reduce the client cost by up to two orders of magnitude, e.g., 277$\times$ for computation time and 281$\times$ for bandwidth usage.
\sys{} also provides flexible trade-offs between the generated model accuracy and the client cost.

In summary, the main contributions of this paper are:
\begin{itemize}
	\item We propose federated neural architecture search, a novel paradigm to automate the generation of DNN models on decentralized data.
	\item We present \sys, a practical framework that enables efficient federated NAS. The core of \sys{} is to fully leverage the insufficient candidate tuning, an intrinsic NAS characteristic, and incorporate key optimizations to reduce on-client overhead.
	%by parallell tuning, early dropping candidates, and dynamic training degree.
	\item We evaluate \sys with extensive experiments. Results show that \sys{} is able to generate a sequence of models under different resource budgets with as high accuracy as traditional NAS algorithm without centralized data, and  significantly reduce computational and communication cost on clients compared to straightforward federated NAS designs.
\end{itemize}

\section{related work}

\noindent \textbf{Neural architecture search (NAS)}
%Automated Machine Learning (AutoML) is a general concept aiming to relieve the developers from the end-to-end operational stages of machine learning, including data pre-processing~\cite{katara}, model selection~\cite{feurer2015efficient}, hyperparameter tuning~\cite{autoweka}, etc.
%Among the many sub-fields, neural architecture search (NAS) has become one of the most interesting and challenging ones.
Designing neural networks is a labor-intensive process that requires a large amount of trial and error by experts.
To address this problem, there is growing interest in automating the search for good neural network architectures.
Originally, NAS is mostly designed to find the single most accurate architecture within a large search space, without regard for the model performance (e.g., size and computations)~\cite{liu2018progressive,rl-nas}.
In recent years, with more attention on deploying neural networks on heterogeneous platforms, researchers have been developing NAS algorithms~\cite{tan2019mnasnet,morphnet,netadapt,wu2019fbnet} to automate model simplifications.
The goal is to generate a sequence of simplified models from an expensive one with the best accuracy under corresponding resource budgets,
i.e., the \textit{pareto frontier of accuracy-computation trade-off}.
 \sys{} is motivated and based on those prior efforts.

%Note that though in this work we focus on only federating NAS, the idea of federation can be applied to the whole AutoML process.
%We leave it as our future work.

\noindent \textbf{Accelerating NAS}
Despite the remarkable results, conventional NAS algorithms are prohibitively computation-intensive.
The main bottleneck is the training of a large number of model candidates, which often takes up to thousands of GPU hours~\cite{zoph2018learning}.
As a trade-off, many NAS algorithms~\cite{netadapt,liu2018progressive,cai2018path,tan2019mnasnet,morphnet} propose to search for building blocks on proxy tasks, such as training for fewer epochs, starting with a smaller dataset, or learning with fewer blocks.
Our work also utilizes and retrofits such proxy tasks as insufficient tuning during the search process.
Recent work has explored weight sharing across models through a hypernetwork~\cite{brock2017smash,pham2018efficient} or an over-parameterized one-shot model~\cite{conf/icml/BenderKZVL18,cai2018proxylessnas} to amortize the cost of training.
Those methods, however, target at generating only one model and often break the high parallelism of NAS,
making them not suitable to our target scenario, i.e., generate multiple models under different resource budgets in federated settings.
A few efforts have proposed distributed systems~\cite{ATM,Vizier,Katib} for automated machine learning tasks.
%ATM~\cite{ATM} is a distributed AutoML platform that works in a load-balanced fashion to quickly deliver results in the form of ready-to-predict models, confusion matrices, cross-validation results, and training timings.
%Google Vizier~\cite{Vizier} is a Google-internal service for performing black-box optimization in machine learning with high scalability.
%Katib~\cite{Katib} is a general AutoML platform that abstracts AutoML algorithms into functionally isolated components and containerize each component as a micro-service.
%Zoph et al~\cite{rl-nas} uses a parameter server to speed up the NAS process by replicating the model parameters among multiple machines.
Such work assume the training data is centralized on cloud instead of decentralized on clients.
As a comparison, we face some unique challenges: data distributions are non-iid and highly-skewed~\cite{yang2021characterizing,cai2022autofednlp}, client devices are resource-constrained~\cite{xu2022mandheling,wang2022melon,zhang2022comprehensive,cai2021towards}, etc.

\noindent \textbf{Federated learning (FL)}
~\cite{mcmahan2016communication} is a distributed machine learning approach to enabling the training on a large corpus of decentralized data residing on devices like smartphones.
By decentralization, FL addresses the fundamental problems of privacy, ownership, and data locality.
Though our proposed \sys{} approach borrows the spirits from FL, all existing FL research focus on training one specific model instead of the end-to-end procedure of automatic architecture search.
As a result, their designs miss important optimizations from intrinsic characteristics of NAS such as early dropping out model candidates (more details in Section~\ref{sec:back:opt}).
Further enhancements to FL, e.g.,
improving the privacy guarantees by differential privacy~\cite{fl-dp} and secure aggregation~\cite{fl-secure},
reducing the communication cost among cloud and clients through weights compression~\cite{konevcny2016federated,mcmahan2016communication},
are complementary and orthogonal to \sys.

\begin{comment}
Enhancements have been innovated for federated learning.
For example, researchers have been exploring how to further improve the privacy guarantees for FL by differential privacy~\cite{fl-dp} and secure aggregation~\cite{fl-secure}.
Some work aim to reduce the communication cost among cloud and clients through weights compression~\cite{konevcny2016federated,mcmahan2016communication}.
MOCHA~\cite{smith2017federated} enables federated multi-tasking learning through a novel systems-aware optimization.
\cite{bonawitz2019towards} shares lessons in building real-world FL applications from the system perspective.
All those work are orthogonal and compatible with \sys.
\end{comment}

\noindent \textbf{Federated AutoML} The initial concept of automating machine learning with decentralized data is first proposed in~\cite{kairouz2019advances}.
\sys{} is one of the earliest frameworks that fulfill this concept in a practical way.
A concurrent work with \sys{}, FedNAS~\cite{fednas}, also applies federated learning on NAS.
However, FedNAS targets at cross-organization scenario (cross-silo) where each client is a GPU-equipped edge server located in an organization,
while \sys{} targets at cross-device scenario where each client is a mobile device with much less hardware resources.
As a consequence, the methodology also differs: FedNAS treats each client as a local searcher while \sys{} treats a group of many clients as a trainer for a specific candidate.
\section{methodology}
We define our target problem and identify the challenges. 

\subsection{Problem Statement}\label{sec:back:prob}
\noindent \textbf{Objective}
Intuitively, the goal of our proposed federated NAS framework is to provide optimal models to run on mobile devices in an \textit{automatic} and \textit{privacy-preserving} way.
For automation, the framework can begin with a well-known network architecture, e.g., \textit{MobileNet}, and generate a sequence of simplified models under different resource budgets without any developers' manual efforts~\cite{liu2014imashup}.
To preserving privacy , the framework requires no training data (e.g., input corpus, images) to be uploaded to a centralized cloud or shared among devices. Such application scenarios are abounding: next-word prediction~\cite{fl-dp,hard2018federated},  speech keyword spotting~\cite{leroy2019federated}, image classification~\cite{liu2018secure}, etc.
As traditional NAS frameworks do, the goal of federated NAS can be formulated as following:
\begin{align*}
\argmax_{DNN} \quad & Acc(DNN) \\
\textit{subject to} \quad & Res_{j}(DNN) \le Bud_{j}, j = 1,2...n
\end{align*}
where $DNN$ is a simplified NN model,
$Acc(·)$ computes the accuracy,
$Res_i (·)$ evaluates the resource consumption of the $i^{th}$ resource type,
and $Bud_i$ is the budget of the $i^{th}$ resource and the constraint on the optimization.
The resource type can be computational cost (MACs), latency, energy, memory footprint, etc., or a combination of these metrics.
The main terminologies and symbols used in this work are summarized in Table~\ref{tab:term}.
For simplicity, we only consider one resource type in this work (i.e., $n$=1).

% !TeX root = main.tex

\begin{table}[]
\footnotesize
\begin{tabular}{|L{1.2cm}|L{6.5cm}|}
\hline
\textbf{Notation} & \textbf{Definitions or Descriptions} \\ \hline \hline
iteration & Cloud loops for different decayed resource budgets ($T$) \\ \hline
round & Cloud loops for fusing the gradients from different clients ($R$) \\ \hline
epoch & Client loops for training on local dataset each round ($E$) \\ \hline
short-term fine-tune & Insufficient re-training of model candidates during neural architecture search without convergence \\ \hline
long-term fine-tune & Sufficient re-training at the end of whole search process \\ \hline
$GM$ & Global model maintained by cloud that achieves best accuracy under certain resource budget \\ \hline
$PM$ & DNN candidate that is simplified from a $GM$ \\ \hline
%$Res(M)$ & The resource consumption of model $M$\\ \hline
\end{tabular}
\caption{Terminologies and symbols}
\label{tab:term}
\vspace{-0.8cm}
\end{table}

\noindent \textbf{Contributors}
A typical federated setting assumes that there are substantial distributed devices available for training, e.g., tens of thousands~\cite{bonawitz2019towards}.
A device can be a smartphone, a tablet, or even an IoT gadget depending on the target scenario.
Each device contains a small number of data samples locally, and limited hardware resources (e.g., computational capacity and network bandwidth).
%The difference between \sys{} and prior NAS frameworks~\cite{netadapt} is that the training data is distributed among client devices.
%During the whole process,  only the model weights can be uploaded to the cloud but the data stays on each client.

\subsection{Federating State-of-the-Art NAS Algorithm}\label{sec:back:skeleton}
Intuitively, any NAS algorithm can be leveraged to work on decentralized data.
We base our approach on one of the state-of-the-art: NetAdapt~\cite{netadapt}.
Besides its superior performance as reported, it has another advantage:
NetAdapt generates multiple DNN candidates for each iteration, and selects one of them based on their performance.
Those candidates can be trained and tested in parallel without any dependency. Indeed, this suits well into the federated setting where lots of devices run independently.

% !TeX root = main.tex

\begin{figure}[t]
	\centering
	\includegraphics[width=0.48\textwidth]{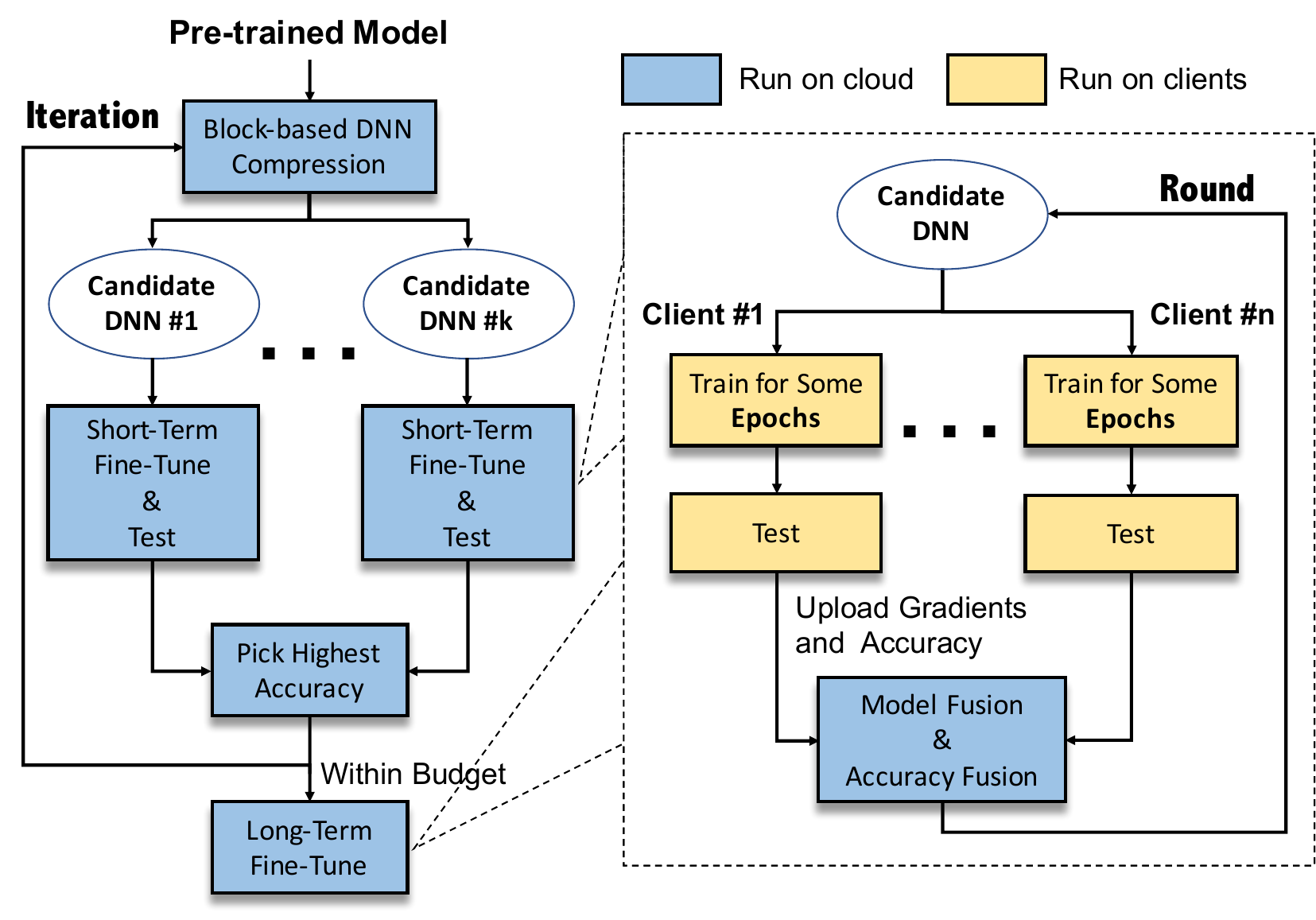}
	%\vspace*{-.6cm}
	\caption{Simplified workflow of \sys. The left part (outside of the dash box) also represents the workflow of original NetAdapt~\cite{netadapt}, the basis of \sys.}
	\label{fig:workflow}
\end{figure}
Figure~\ref{fig:workflow} shows the workflow of NetAdapt (with only left part of the figure) and its federated version (the whole figure).
Generally speaking, NetAdapt iterates over monotonically decreasing resources budgets, for each of which it generates multiple compressed DNN candidates, fine-tunes each candidate, and picks the optimal one with highest accuracy. Finally it performs a long-term fine-tune on the optimal models to convergence.
To enable NetAdapt to run on decentralized data, i.e., under federated settings, we can simply replace the training (both short-term and long-term fine-tune) and testing part with a FL-like process, in which a model will be trained at available clients and fused at cloud for many rounds.
Section~\ref{sec:design} will present more details of how \sys{} works.

%The reasons are threefolded:
%(1) NetAdapt starts with a pre-trained model, and keeps pruning the model in an iterative, resource-aware manner.
%It fits well into the problem we intend to solve: generating optimal models under different resource budgets.
%(2) NetAdapt is heuristic-based, thus runs much faster than learning-based approaches.
%For example, on CIFAR10 dataset, NetAdapt runs only \note{X} hrs while XXX runs X hrs.   
%(3) NetAdapt generates multiple pruning candidates for each iteration, and selects one of them based on their performance.
%Those candidates can be trained and tested parallelly without any dependency, which suits well into the federated setting where lots of devices can run at the same time independently.

\subsection{Challenges and Key Optimizations}\label{sec:back:opt}
The major challenge of federated NAS is the heavy on-client computational and communication cost.
Consequently, the end-to-end process of federated NAS can be excessively time-consuming.
Taking communication cost for short-term fine-tune as an example, the total uplink bandwidth usage can be roughly estimated as
\[\sum_{i}\sum_{j}(grad\_size(DNN_{i,j}) \cdot client\_num(DNN_{i,j}) \cdot round\_num)\]
where $grad\_size()$ calculates the model gradient size, $client\_num()$ is the number of clients involving training $DNN_{i,j}$,
$i$ and $j$ iterate over all resource budgets (depending on the developer configurations) and DNN candidates (depending on the model architecture), respectively.
Communication cost is known to be a major bottleneck in federated learning~\cite{mcmahan2016communication}, and it will be further amplified by the large number of DNN candidates and resource budgets to be explored during NAS.
%Indeed, how to reduce the client cost in federated learning has drawn a lot of research attention recently~\cite{konevcny2016federated,mcmahan2016communication}.
%Those efforts, however, focus on optimizing the single model training, thus miss the opportunities from intrinsic characteristics in NAS.

We identify the key opportunity as \textit{how sufficient shall each DNN candidate be tuned during the search process}.
While some work realized the tuning can be short-term without getting converged, but they did not explore how long such a process is sufficient.
By our study, we find that the tuning of each DNN candidate can be parallel (across clients), dynamic (across time), heterogeneous (across models).
In the following sections, we introduce three key optimizations provided by \sys, where the first one is to reduce $client\_num$,  and the other two are to reduce $round\_num$.

\noindent \textbf{Training candidates on partial clients in parallel}
One opportunity of speeding up federated NAS comes from the huge amount of client devices that can participate in the training process.
In common FL setting, a device is available when it is idle, charged,  under unmetered network (e.g., WiFi), and so on.
As reported by Google~\cite{bonawitz2019towards}, tens of thousands of devices are available for FL at the same time.
However, only hundreds of devices can be efficiently utilized in parallel due to the limitation of the state-of-the-art gradients fusion algorithm (e.g., \texttt{FedAvg}).
By training and testing DNN candidates on separated groups of clients, we not only reduce the average computational and communication cost of each candidate,
but also scale out better with the large number of available clients.
%To further scale out with the number of available clients, we observe that training and testing each candidate doesn't require all training data (or clients) to be involved.
%Instead, only a small number of data is enough to fine-tune and pick the optimal candidate~\cite{netadapt}.
%It indicates that not only one model candidate can be parallelly trained on multiple clients, but different candidates can be parallelly trained on different sets of clients.
It motivates us to organize all clients into a two-level hierarchy for high parallelism (more details in Section~\ref{sec:design:schedule}).

% !TeX root = main.tex

\begin{figure}[t]
	\centering
	\begin{minipage}[b]{0.24\textwidth}
	\includegraphics[width=1.0\textwidth]{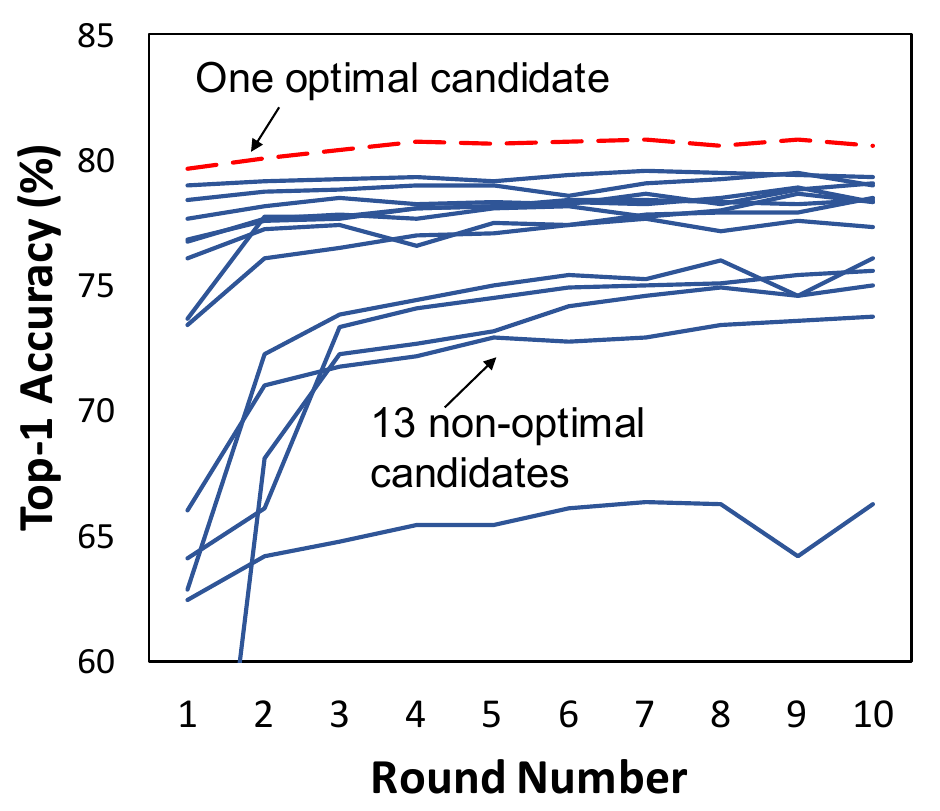}
	%\vspace*{-.6cm}
	\subcaption{2nd Iteration}
	\end{minipage}
	~
	\begin{minipage}[b]{0.24\textwidth}
	\includegraphics[width=1.0\textwidth]{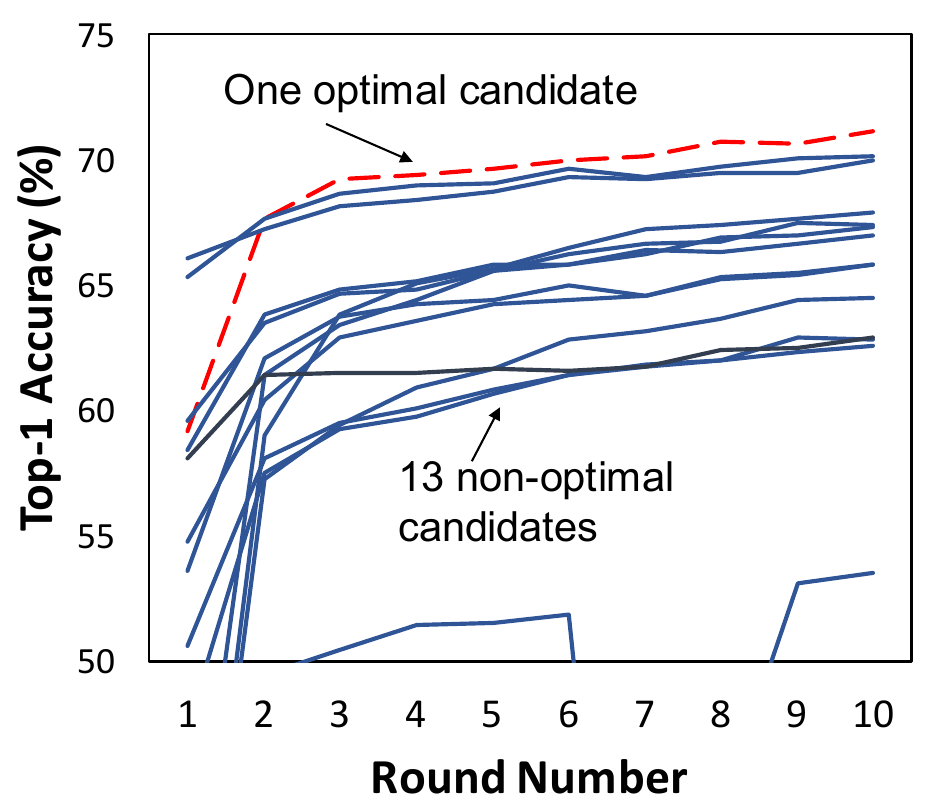}
	%\vspace*{-.6cm}
	\subcaption{8th Iteration}
	\end{minipage}
	\caption{The accuracy of each DNN candidate as training goes on with more rounds.
		Each line represents one DNN candidate, and the red dashed one is the optimal one that shall be picked.
		Dataset: ImageNet, model: MobileNet.}
	\label{fig:drop}
\end{figure}

\noindent \textbf{Dynamic round number}
We studied the performance of different DNN candidates at different resource iterations.
As illustrated in Figure~\ref{fig:drop}, each line represents the accuracy (y-axis) of a candidate with different training rounds (x-axis), and the red dashed one is the optimal candidate to be picked as it achieves the highest accuracy after all rounds of training done.
By comparing the two subfigures, we find that at early iterations the DNN candidates, especially those with higher accuracy, reach stable condition much earlier than later iterations.
This is because our algorithm starts with a pre-trained model: as it proceeds the impacts from inefficient tuning accumulate and the model parameters become more and more random.
Such insight motivates us towards using dynamic round numbers, e.g., a smaller one for early iterations and keep increasing the number in later stages.
While the round number becomes larger, it is worth noting that the model complexity ($grad\_size(DNN)$) decreases as the algorithm proceeds.
It makes the optimization quite effective in reducing clients' overheads.

\noindent \textbf{Early dropping out non-optimal candidates}\label{sec:back:drop}
Figure~\ref{fig:drop} also shows that the optimal candidate (the red dashed line) quickly outperforms others within 1$\sim$3 rounds.
It guides us to another optimization: early dropping the candidates while only keeping the optimal one being trained with more rounds.
Noting that though optimal candidate has been already picked within several rounds, it still needs to go through more rounds of training.
Otherwise the model accuracy will quickly drop to very low and thus misleading the candidate selection afterwards,
as confirmed by our experiments in Section~\ref{sec:eval:abla}.

In next section we will introduce the details of our federated NAS framework, \sys, which incorporates the aforementioned optimization techniques.
\section{the \sys{} system}\label{sec:design}

\noindent \textbf{Overview}
The pseudo code of {\sys}'s workflow is shown in Algorithm~\ref{algo:workflow}.
\sys{} maintains a model called \textit{global model} ($GM$) among cloud and clients, which starts with an expensive one and will be iteratively adapted until it meets the required resource budget.
The network architecture of the initial global model ($GM_0$) is given by the developers, e.g., MobileNet.
It can be either a pre-trained model or actively trained through federated learning as part of \sys.
The goal of each iteration (line 2--11) is to adapt the global model to a smaller one through the cooperation between cloud and clients, i.e.,
under the budget of $Res(GM_t)-\Delta R_t$ where $\Delta R_t$ indicates how much the constraint tightens for the $t^{th}$ iteration (a similar concept of learning rate) and can vary from iteration to iteration.
The algorithm terminates when the final resource budget is satisfied.
\sys{} outputs the final adapted model and also generates a sequence of simplified models at intermediate iterations (i.e., the highest accuracy network picked at each iteration $\langle GM_1$, ..., $GM_T\rangle$) that form the efficient frontier of accuracy to resource trade-offs.

The $t^{th}$ iteration begins with generating a set of pruned models ($PMs$) as candidates based on $GM_{t-1}$ ($\S$\ref{sec:design:prune}).
Each $PM$ will be scheduled to a group of clients ($\S$\ref{sec:design:schedule}),
on which the model will be repeatedly i) downloaded to each client within the group (line 14--24);
ii) trained and tested via the local dataset on that client (line 34--38); iii) collected to cloud and fused into a new model for many rounds (line 27--32).
During this process, all $PMs$ except the optimal one will dropped out (line 25--26, $\S$\ref{sec:design:drop}).
This picked $PM$ represents the most accurate model under the current resource budget, thus making it the next global model $GM_t$ (line 10).
%The testing results from all the devices within the same group will be fused to an accuracy, which indicates the quality of the $PM$ ($\S$\ref{sec:design:acc-fuse}).
%After all $PM$s have been scheduled and tested, the cloud looks at all the fused accuracy and picks the optimal $PM_{opt}$.
%All devices within the group that train the $PM_{opt}$ will upload their local-tuned model to cloud, where they are fused to a new global model $GM_i$ ($\S$\ref{sec:design:model-fuse}).
Finally, the cloud performs federated learning on the $GM_T$ or other $GMs$ as specified by developers till convergence (line 12, $\S$\ref{sec:design:fl-tune}).

% !TeX root = main.tex

\let\oldnl\nl% Store \nl in \oldnl
\newcommand{\nonl}{\renewcommand{\nl}{\let\nl\oldnl}}% Remove line number for one line

\begin{algorithm}
\caption{The proposed \sys{} framework}\label{algo:workflow}

\small
\SetKwInOut{Input}{input}
\SetKwInOut{Output}{output}
\SetKw{KwBy}{by}

\Input{the init model $GM_0$, the resource budget $R$, iteration number $T$, candidate drop ratio $\alpha\%$
}
\Output{a list of models ($GM_1, ... GM_T$) under different resource budgets ($R_1, ..., R_T$ where $R_T=R$)}

\SetKwProg{Fn}{Function}{:}{}

\Fn(\tcp*[h]{run on cloud}){Cloud\_Operation()}{
	\For{$t\gets1$ \KwTo $T$}{
		$R_t \gets$ next resource budget \\
		$round\_num \gets$ round number at current iteration \\
		$groups \gets$ partition all available clients into groups \\
		$PMs \gets$ generate a list of simplified models by pruning different layers of $GM_{t-1}$ \\
		\For{$k\gets1$ \KwTo $round\_num$}{
			$PMs \gets$ \textit{Cloud\_One\_Round}($PMs, groups$) \\
		}
		$GM_t \gets PMs[0]$
	}
	perform FL on $GM_T$ or other $GMs$ as user specify \\
}

\Fn{Cloud\_One\_Round($PMs$, $groups$)}{
	\For{each $PM_i$ in $PMs$ \textbf{in parallel}} {
		$G_j \gets$ wait to get a free group from $groups$ \\
		lock($G_j$) \\
		\For{each client $C$ in $G_j$ \textbf{in parallel}}{
			send $PM_i$ to $C$ \\
			invoke $C.$\textit{Client\_Operation}($PM_i$) \\
			collect $acc, test\_num$ from $C$ \\
		}
		$PM_i\_acc \gets$ fuse all collected $acc$ based on $test\_num$ \\
		unlock($G_j$) \\
	}
	$drop\_num \gets max(\alpha\% \cdot PMs.len(), PMs.len()-1)$ \\
	remove $drop\_num$ models with lowest $PM\_acc$ from $PMs$ \\
	\For{each $PM_i$ in $PMs$ \textbf{in parallel}}{
		$G_j \gets$ the group that runs $PM_i$ \\
		collect $grad, train\_num$ from all clients within $G_j$ \\
		$grad \gets$ fuse all $grad$ based on $train\_num$ \\
		$PM_i \gets$ $PM_i$ + $grad$
	}
	\KwRet $PMs$
}

\Fn(\tcp*[h]{run on remote clients}){Client\_Operation($Model$)}{
	split local dataset into training and validation set \\
	$grad \gets$ train $Model$ on local training dataset for $E$ epoches \\
	$acc \gets$ test $Model$ on local validation dataset \\
	store $grad$, $acc$, $train\_num$, $test\_num$ locally
}

\end{algorithm}
\vspace{-5pt}

\subsection{Model Pruning}\label{sec:design:prune}
\sys{} adapts a $GM$ based on standard pruning approaches.
More specifically, \sys{} reduces the number of filters in a single CONV (convolutional) or FC (fully-connected) layers to meet the resource budget of current iteration,
as CONV and FC are known to be the computationally dominant layers in most NN architectures~\cite{deepcache}.
To choose which filters to prune, \sys{} computes the $\ell$2-norm magnitude of each filter and the one with smallest value will be pruned first.
More advanced methods can be adopted to replace the magnitude-based method, such as removing the filters based on their joint influence on the feature maps~\cite{yang2017designing}.

By adapting, \sys{} generates $K$ pruned model candidates $PM$s, where $K$ equals to the sum of CONV and FC layer numbers, e.g., 14 for MobileNet.
For larger models, we can also speed up the adaptation process by treating a group of multiple layers as a single unit (instead of a single layer), e.g., residual block in ResNet~\cite{resnet}.

\subsection{Clients Partitioning \& Scheduling}\label{sec:design:schedule}
The goal of this stage is to partition the clients that are available for training into different groups.
Each group contains one or multiple clients, and the number of groups ($\mathcal{K}$) is given by developers.
The partition starts once the cloud determines which clients are available and their associated information (i.e., data number and data distribution, see below) has been uploaded.
Since the availability of clients is dynamic depending on the user behavior and device status, the partition needs to be performed at each iteration.
Each $PM$ will be scheduled to one group for training (i.e., short-term fine-tune) and testing.

\noindent \textbf{How to partition}
A good partition follows two principles.
First, the total data number of each group shall be close and balanced.
This is to ensure that each $PM$ is tuned and tested on enough data to make the results trustworthy, and also ensure high parallelism without being bottlenecked by large groups.
Second, the data distribution of each group shall be representative of the dataset from all clients.
Since in federated setting the data owned by each client is often non-iid, a random partition may lead to groups with biased data and makes the resultant accuracy non-representative.
In such a case, our algorithm may choose the wrong candidate.

To formulize the two policies above, we denote a partition $\euscr{P}$ as $\{G_1, ..., G_{\mathcal{K}}\}$, and the total data number within $G_i$ as $d_i$ which is simply summed over the data number of all clients within $G_i$.
\begin{align*}
\argmin{\euscr{P}} \quad & \frac{1}{\mathcal{K}} \sum_{i=1,...,\mathcal{K}}{dist\_dist(G_i, G_{all})} \\
\textit{subject to} \quad & \max(d_j) \le \textit{r} \cdot \min(d_j), \quad j = 1,...,\mathcal{K}
\end{align*}

Here, $dist\_dist()$ calculates the distance between the data distributions of two groups, $G_{all}$ is an imaginary group including the data from all clients,
\textit{r} is a configurable variable that controls how unbalanced \sys{} can tolerate about the data sizes across different groups (default: 1.1).
This equation can be approximately solved by a greedy algorithm: first sorting all clients by their data number, then iteratively dispatching the largest one to a group so that the data size balance is maintained (i.e., the inequality) while the smallest average distribution distance is achieved.

For classification tasks, which is the focus of this work, \sys{} uses the normalized number of each class type to represent the data distribution, i.e., a vector 
$v=(v_1,v_2,...,v_m)$ where $v_i$ equals to the ratio of data numbers labeled with $i^{th}$ class type.
The distribution distance is computed as the Manhattan distance between such two vectors.
Note that the ratio of different class types can be considered to be less privacy-sensitive compared to the gradients that need to be uploaded for many times, so it shall not compromise the original privacy level of federated setting.
Nevertheless, the distribution vectors can be further encrypted through secure multiparty computation~\cite{lindell2005secure}.

\noindent \textbf{How to schedule}
Each $PM$ will be scheduled to a random group for training and testing.
If all groups are busy, cloud will wait until one has finished and schedule the next $PM$ to this group.

As an important configuration to be set by the developers, the number of groups ($\mathcal{K}$) makes the trade-offs between the quality of neural architecture selection and the computational cost imposed on client devices.
A larger $\mathcal{K}$ promises higher parallelism so that the NAS process can be faster, but also means the training and testing data provisioned to each $PM$ is less.
Our experiments in Section~\ref{sec:eval} will dig into such trade-offs and provide useful insights to developers in determining a proper group number.

\subsection{Candidate Dropping and Selection}\label{sec:design:drop}
\noindent \textbf{Short-term fine-tune on decentralized data} Each $PM$ will be trained and tested on the scheduled group for many rounds, similar to the methodology of federated learning.
At each round, every client within the group downloads the newest $PM$ version, then trains (\textit{local-tune}) and tests the model.
The training and testing datasets are both split from the client's local dataset.
The local-tune takes multiple epochs ($E$) to reduce round number and communication cost~\cite{mcmahan2016communication}.
The training and testing results, i.e., gradients and accuracy, associated with the dataset size, will be uploaded to cloud.
The gradients will be fused to update the model candidate $PM$ on cloud, and the accuracy will be fused as the metric to pick the optimal model candidate after all rounds.

Guided by our finding in Section~\ref{sec:back:drop}, \sys{} reduces the on-client computational and communication cost during short-term fine-tune process through dynamic round number and early dropping candidates.
More specifically, \sys{} increasingly trains each candidate with more rounds as iterations go on.
For each round, \sys{} collects the local accuracy from clients and fuse them into a weighted accuracy for each $PM$.
The $\alpha\%$ ones with largest accuracy degradation (defined below) will be dropped and no longer tuned.
For the rest of the valid candidates, \sys{} collects the gradients from clients and fuses them into a new $PM$.
As round goes on, fewer and fewer candidates need to be tuned and tested.
%After $R$ rounds, \sys{} selects the $PM$ with highest accuracy among all left candidates, and treats it as the next $GM$.
Noting that the accuracy is fused first so that the gradients of the dropped candidates at this round do not need to be uploaded.

The goal of this short-term fine-tune is to regain accuracy of $PMs$.
This step is important while adapting small networks with a large resource reduction because otherwise the accuracy will drop to zero, which can cause \sys{} to choose the wrong model candidate.
One main difference between this stage and a standard FL process is that this stage takes relatively smaller number of iterations (i.e., short-term) without requiring the model to converge.

\noindent \textbf{Accuracy fusion and comparison}
The accuracy generated by each client will be uploaded to the cloud.
For a given $PM_i$ and its scheduled group $G_j$, once the cloud receives all accuracy of the clients within the same group $G_j$, it combines the accuracy into a new one by weighting the testing data numbers on the same client:
\[
PM_i\_acc =\frac{\sum_{s}{(test\_num_{j,s} \cdot acc_{j,s})}}{\sum_{s}{test\_num_{j,s}}}, \quad s=1...g_j
\]
where $test\_num_{j,s}$ and $acc_{j,s}$ are the testing data number and testing accuracy reported by the $s^{th}$ client of $j^{th}$ group correspondingly,
$g_j$ is the client number of $j^{th}$ group.

With the accuracy of all model candidates computed at each round, \sys{} drops the models with largest accuracy degradation.
Note that each $PM$ may have different resource consumptions (Section~\ref{sec:design:prune}), we use the ratio of accuracy degradation to the resource consumption reduction over the previous $GM$ (i.e., the unpruned model at the beginning of this iteration):
\[
acc\_degradation = \frac{Acc(prev\_GM) - PM_i\_acc}{Res(prev\_GM) - Res(PM_i)}
\]

\noindent \textbf{Model fusion}\label{sec:design:model-fuse}
For a given $PM_i$ and its scheduled group $G_j$, \sys{} fuses the gradients from all clients within $G_j$ by weighting the training data numbers used in local-tune.
\[
fused\_grad_i = \frac{\sum_{s}{(train\_num_{j,s} \cdot grad_{j,s})}}{\sum_{s}{train\_num_{j,s}}}, \quad s=1...g_j
\]
\[
PM_i \gets PM_i + fused\_grad_i
\]
where $train\_num_{j,s}$ and $grad_{j,s}$ are the training data number and gradients uploaded from the $s^{th}$ client of $j^{th}$ group correspondingly.

%Note that only the clients within the group that generates the most accurate $PM$ will upload the models, while other clients only upload some numerical results like accuracy and data number.
%Such design significantly reduces the communication overhead between clients and cloud as the models can be up to tens of MBs.

\subsection{FL-tuning}\label{sec:design:fl-tune}
Once \sys{} finishes the model search process above, a sequence of models have been generated, i.e., $GM_1, ..., GM_T$.
As the final stage, \sys{} performs a standard federated learning on $GM_T$ or other $GMs$ (called \textit{FL-tune}) if needed by the developer.
The goal of this stage is to make the obtained models converge.
\sys{} can utilize any existing FL algorithm to run FL-tune and currently it uses one of the state-of-the-art $FedAvg$~\cite{mcmahan2016communication}.
When multiple $GMs$ are demanded, \sys{} can still utilize the partitioned clients to train them in parallel.

%After the fused model, i.e., a new global model, has been generated, \sys{} decides whether to perform a FL on the model (\textit{FL-tune}).
%The goal of FL-tune is to regain the model accuracy and avoid that the accumulative accuracy degradation after many iterations will lead to wrong model candidates selection.
%Such degradation may come from the reduced model parameters, unrepresentative group data, etc.
%
%To reduce the high computational overhead of FL-tune, \sys{} only performs a FL-tune only when it monitors a large accuracy degradation between two consecutive global models, i.e., \[Acc(GM_i) - Acc(GM_{i-1}) > \theta\]
%Here $\theta$ is a user-specified threshold that controls when to perform FL-tune.
%\sys{} can utilize any existing FL algorithm to run FL-tune and currently it uses one of the state-of-the-art $FedAvg$~\cite{mcmahan2016communication}.
\section{evaluation}\label{sec:eval}

In our experiments, we mainly evaluate three parts of performance:
1) $\S$\ref{sec:eval:acc}: does \sys{} generate high accuracy models under different resource budgets?
2) $\S$\ref{sec:eval:cost}: what's the computational and communication cost of \sys{} on clients?
3) $\S$\ref{sec:eval:abla}: what's the impacts of \sys{}'s key designs?

\subsection{Experiment Settings}

% !TeX root = main.tex

\begin{table}[t]
\footnotesize
\begin{tabular}{lL{1.9cm}L{1.3cm}L{0.9cm}L{1cm}}
\textbf{Dataset} & \textbf{Model} & \textbf{Task} & \textbf{Client number} & \textbf{Data per client} \\ \hline
ImageNet (\textit{iid}) & MobileNet (13 CONV, 1 FC) & Image classification & 1,500 & 915.0 \\ \hline
Celeba (\textit{non-iid}) & Simplified AlexNet (6 CONV, 1 FC) & Face attrs classification & 9,343 & 21.4 \\ \hline
\end{tabular}
\caption{Datasets and models used in experiments.}
\label{tab:dataset}
\vspace{-0.5cm}
\end{table}

\noindent \textbf{Datasets}
As shown in Table~\ref{tab:dataset}, we tested \sys{} on 2 datasets commonly used for federated learning experiments:  ImageNet~\cite{imagenet} (iid) and Celeba~\cite{celeba} (non-iid).
For ImageNet, we randomly split it into 1,500 clients.
For Celeba, we split it into 9,343 clients based on the identities of face images. We re-used the scripts of LEAF~\cite{leaf}, a popular federated learning framework, to pre-process Celeba data and generate non-iid data.
Each Celeba image is tagged with 40 binary attributes. We randomly select 3 of them (\textit{Smiling, Male, Mouth\_Slightly\_Open}) and combine the 3 features into a classification task with 8 classes.
The dataset on each client was further split to three parts: training set used for short-term fine-tune, validation set used to test the accuracy of DNN candidates, testing set used to evaluate the final accuracy of each simplified model (6:2:2).

\noindent \textbf{Models}
We applied \sys{} on two models:
MobileNet~\cite{mobilenet} (for ImageNet, 224x224 input size), a widely used CNN network for mobile applications;
A simplified AlexNet, which we call ConvNet (for Celeba, 128x128 input size) with sequential CONV, Pooling, and final FC layers.
We did not apply \sys{} on larger networks like ResNet or VGG because small and compact networks are more difficult to simplify; these large networks are also seldom deployed on mobile platforms. 

\noindent \textbf{Resource type} We mainly used multiply-accumulate operations (MACs) as the metric to specify resource budgets.
For MobileNet, we reduce the resource budget by 5\% at each iteration with 0.98 decay.
For ConvNet, we reduce the resource budget by 5\% at each iteration with 0.93 decay.

\noindent \textbf{Alternatives} We compare \sys{} with two state-of-the-art automatic network simplification approaches.
Note that both of them are performed on centralized data.
\begin{itemize}
\item \textit{NetAdapt}~\cite{netadapt} is the basis of \sys{}. We directly reused their open code and kept the original parameter setting.
\item \textit{\multipliers}~\cite{mobilenet} are simple but effective approaches to simplify networks. We used Width Multiplier to scale the number of filters by a percentage across all CONV and FC layers.
\end{itemize}

\noindent \textbf{Hardware of Cloud and Clients}
All experiments were carried out on a high-end server with 12$\times$ P100 Tesla GPUs.
To simulate the client-side computation cost, we used DL4J~\cite{DL4J} to obtain the training speed of MobileNet and ConvNet as well as each pruned DNN candidate on Samsung Note 10.
%We can only run very small batch size (i.e., 2) on the smartphone due to tight memory constraint.
The training speed is then plugged into to our experiment platform, as a way to simulate the on-client computation cost.
The communication cost is also simulated by recording the data transmission between cloud process and client processes.
% Besides the processing time, we also simulate the computational cost (FLOPs per device) and communication cost (MBs per device) for \sys{} and the alternatives.

\subsection{Analysis of Accuracy}\label{sec:eval:acc}

% !TeX root = main.tex

\begin{figure}[t]
	\centering
	\begin{minipage}[b]{0.42\textwidth}
	\includegraphics[width=1.0\textwidth]{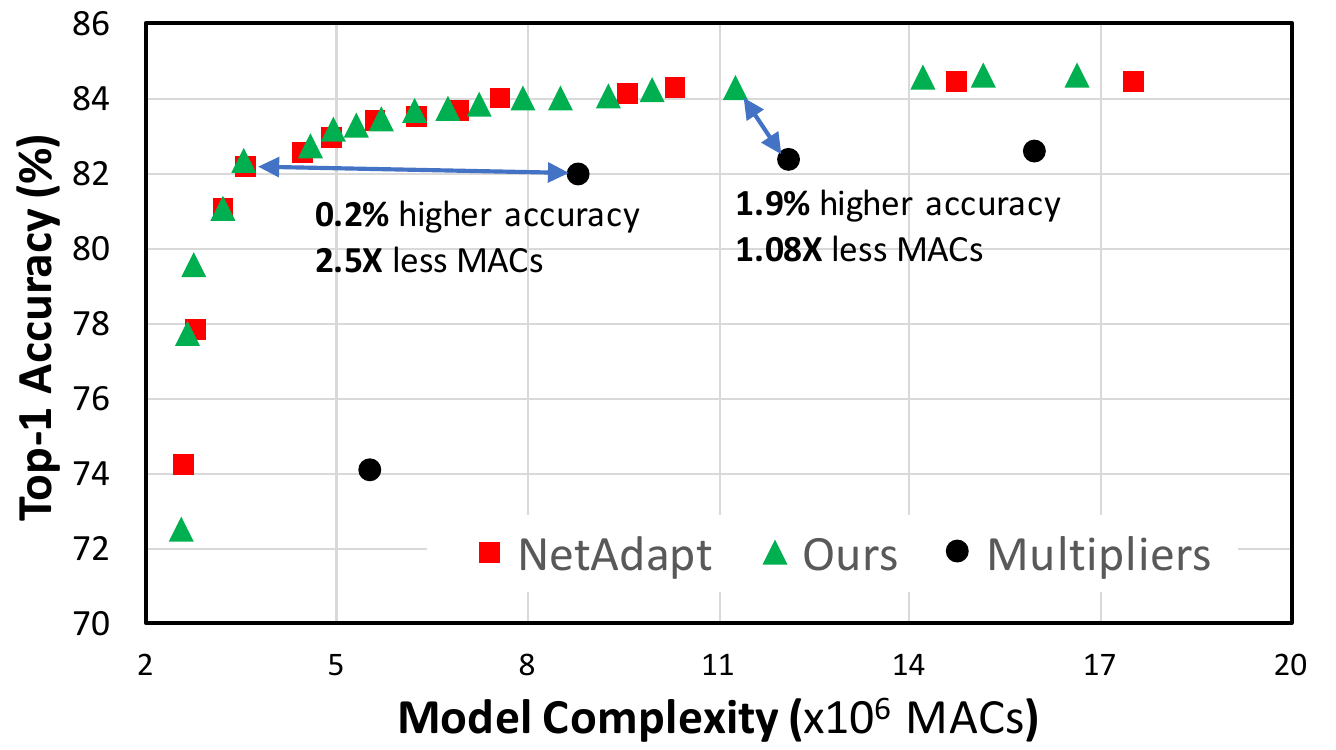}
	%\vspace*{-.6cm}
	\subcaption{Celeba. \sys{} setting: 10 rounds, 4 epochs, 20 groups.}
	\end{minipage}
	\begin{minipage}[b]{0.42\textwidth}
	\includegraphics[width=1.0\textwidth]{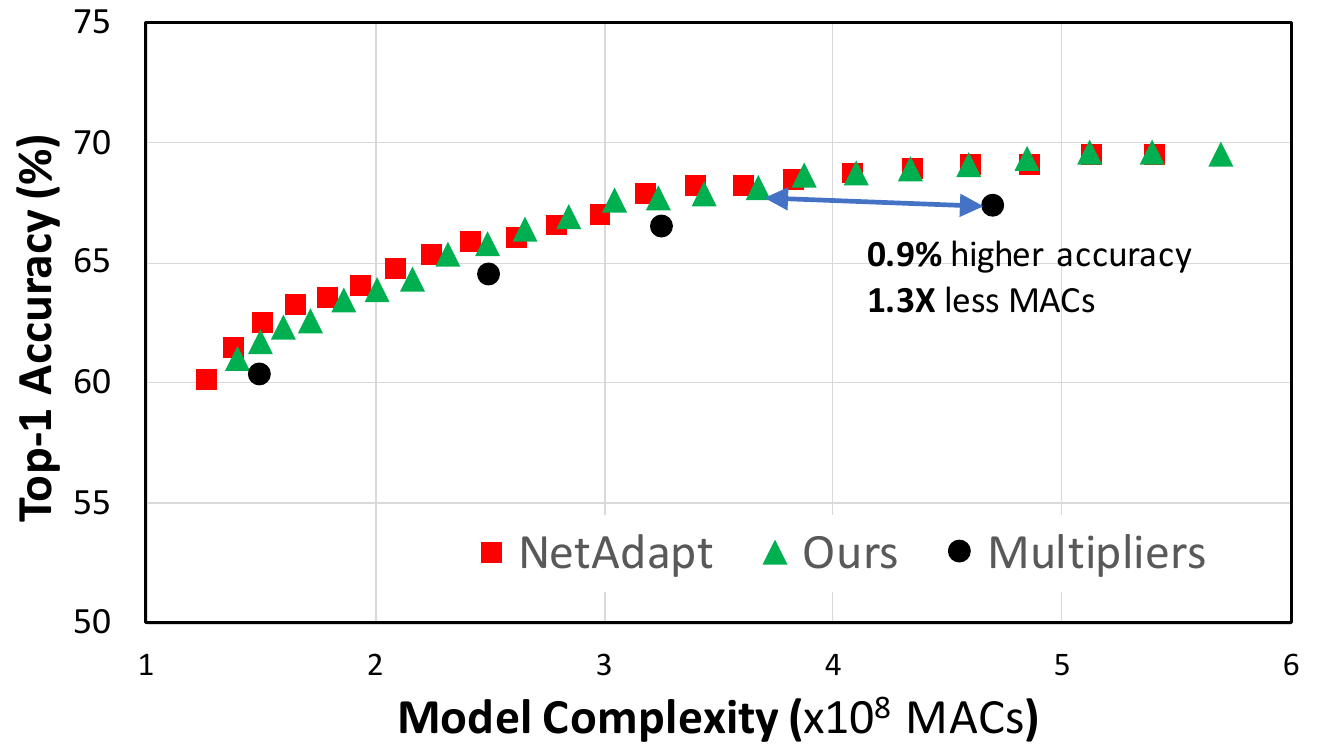}
	%\vspace*{-.6cm}
	\subcaption{ImageNet. \sys{} setting: 20 rounds, 3 epochs, 14 groups.}
	\end{minipage}
	\caption{Accuracy comparison among \sys{} and the alternatives with MACs as the target resource type.}
	\label{fig:perf}
\end{figure}

Figure~\ref{fig:perf} shows the comparison of the models generated by \sys{} and other alternatives.
Overall, \sys{} achieves similar performance as NetAdapt, and both of them significantly outperform \multipliers.
Noting that \sys{} trains models on decentralized data with much better user privacy.
On Celeba, the model generated by \sys{} is up to 2.5$\times$ less complex (specified by MACs) with the similar accuracy or 1.9\% higher accuracy with the same complexity compared to \multipliers.
On ImageNet, the model generated by \sys{} is 1.5$\times$ less complex with 0.8\% higher accuracy compared to \multipliers.

On ImageNet, we notice a performance gap between \sys{} and NetAdapt around 2\% when MobileNet is simplified by more than 70\%.
This is because in our current default setting, the short-term fine-tune is conservative to keep the client cost low, so that sometimes the candidate is not sufficiently trained thus misleading the model selection.
As we will show later, by varying the system configurations (e.g., round number and group number), the accuracy of \sys{} can be further improved to be closer to NetAdapt.

% !TeX root = main.tex

\begin{figure}[t]
	\centering
	\includegraphics[width=0.38\textwidth]{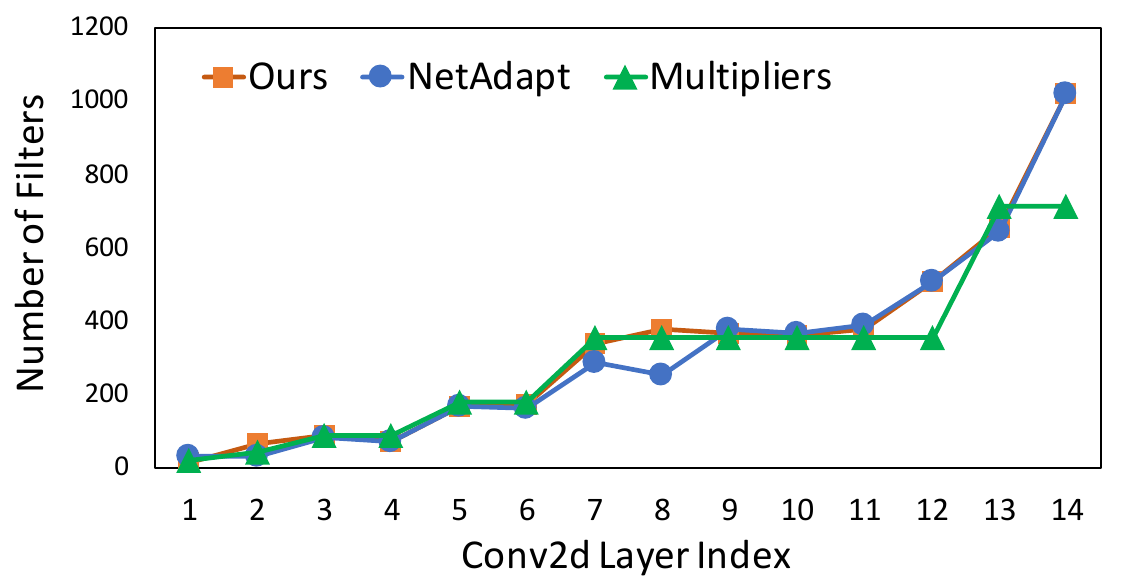}
	%\vspace*{-.6cm}
	\caption{When adapting MobileNet on ImageNet to 50\% MACs, \sys{} and NetAdapt generate similar network architectures, while more different from \multipliers.}
	\label{fig:eval-arch}
\end{figure}
We then studied how the network architectures look like when adapting MobileNet to 50\% MACs on ImageNet using different approaches.
As illustrated in Figure~\ref{fig:eval-arch}, \sys{} generates similar network architecture as NetAdapt but different from \multipliers.
This well explains the performance similarity/gap between \sys{} and the alternatives shown above.

\subsection{Analysis of Client Cost}\label{sec:eval:cost}

We studied how much improvements and trade-offs brought by our key optimizations introduced in Section~\ref{sec:back:opt}.
By default, we drop 33\% candidates after each round, thus all non-optimal candidates will be dropped after 3 rounds.
The dynamic round numbers used are (1-5 iters: 5 rounds; 7-10: 10; 11-15: 15; >15: 20) for ImageNet and (1-5 iters: 2 rounds; 6-10: 5; 11-15: 8; >15: 10) for Celeba.
The group numbers for ImageNet/Celeba are 15 and 20, respectively.
The settings are consistent with the accuracy experiments in Figure~\ref{fig:perf}.
Here we only report the on-client cost for short-term fine-tune during model search, excluding the cost for long-term fine-tune at the last step and the potential federated learning for the initial model. This is because the short-term fine-tune is often more computational intensive, while the latter ones depend on further user specifications, e.g., what models are needed for deployment.

% !TeX root = main.tex

\begin{figure}[t]
	\centering
	\begin{minipage}[b]{0.24\textwidth}
	\includegraphics[width=1.0\textwidth]{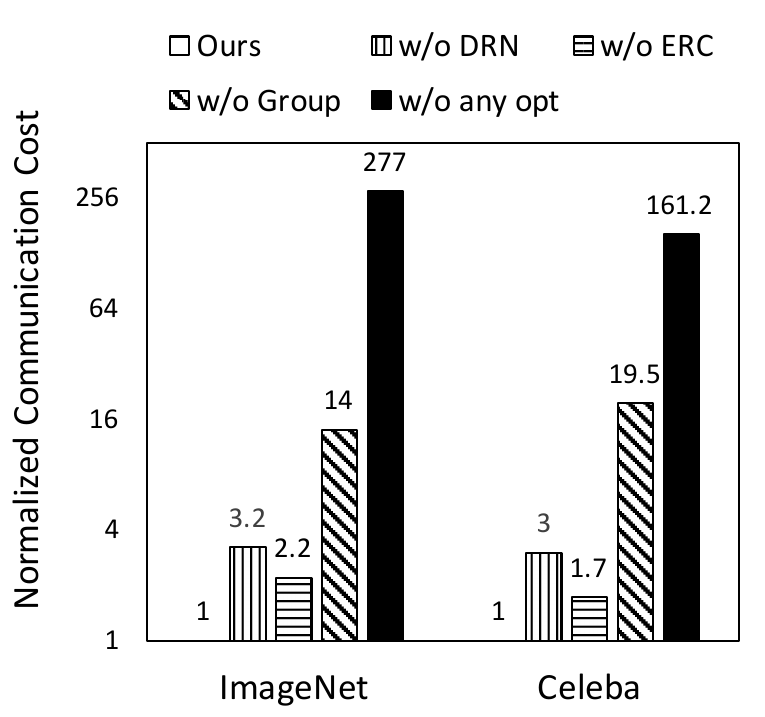}
	%\vspace*{-.6cm}
	\subcaption{Communication cost}
	\end{minipage}
	~
	\begin{minipage}[b]{0.24\textwidth}
	\includegraphics[width=1.0\textwidth]{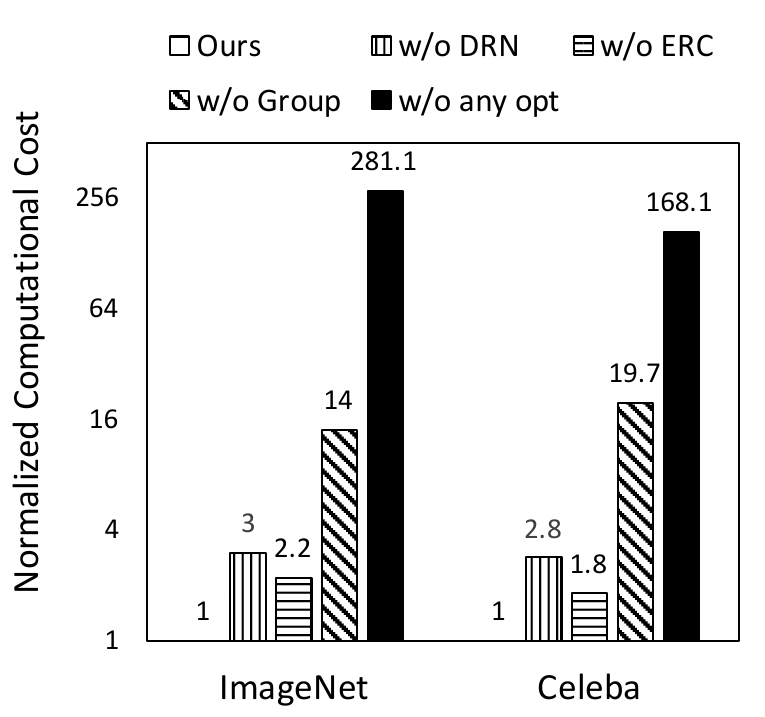}
	%\vspace*{-.6cm}
	\subcaption{Computational cost}
	\end{minipage}
	\caption{The on-client cost reduction brought by our key optimizations (``DRN'': dynamic round number; ``ERC'': early dropping candidates; ``Group'': parallel training).
		The setting is the same as Figure~\ref{fig:perf}. The y-axis is logarithmic.}
	\label{fig:cost}
\end{figure}

\noindent \textbf{Overall improvements}
As shown in Figure~\ref{fig:cost}, all three techniques can significantly reduce the on-client cost, i.e., computational and communication.
In a naive design of federated NAS with all optimizations disabled, the communication and computational cost are 277$\times$ and 281$\times$ more on ImageNet, and 161$\times$ and 162$\times$ more on Celeba, respectively.
With one technique disabled, i.e., dynamic round number / early dropping candidates / group hierarchy, the cost can be up to 3.2$\times$ / 2.2$\times$ / 19.7$\times$ more.
We observe that the first two optimizations aiming at reducing the round number are more effective at ImageNet.
This is because ImageNet task is more complex than Celeba, so the model requires more short-term fine-tuning (round numbers) thus leaves more headroom for optimizations.

Note that, according to the experiments, our optimizations with the default settings have almost \textit{zero} affects at the model accuracy.
In fact, Figure~\ref{fig:perf} shows that \sys{} already achieves the accuracy upper bound defined by NetAdapt.
Next, we studied the trade-offs between accuracy and cost from two optimizations (early drop candidates and group hierarchy) by varying the default settings.

% !TeX root = main.tex

\begin{table}[t]
\footnotesize
\begin{tabular}{|c|L{1.4cm}|L{1.5cm}|L{2.2cm}|}
\hline
\multicolumn{1}{|l|}{\textbf{Model}} & \textbf{Drop ratio each round} & \textbf{Top-1 Accuracy (\%)} & \textbf{Avg uplink cost per client (MBs)} \\ \hline \hline

\multirow{4}{*}{50\% ConvNet} & 0\% (no drop) & 83.8 (0.0) & 59.7 (0.0) \\ \cline{2-4} 
 & \textcolor{red}{33\% (default)} & \textcolor{red}{83.8 (0.0)} & \textcolor{red}{12.8 (-79\%)} \\ \cline{2-4} 
 & 50\% & 83.5 (-0.3) &  10.7 (-82\%) \\ \cline{2-4} 
 & 100\% & 82.1 (-1.7) & 8.5 (-85\%) \\ \hline \hline
 
\multirow{4}{*}{25\% ConvNet} & 0\% (no drop) & 82.3 (0.0) & 138.0 (0.0) \\ \cline{2-4} 
& \textcolor{red}{33\% (default)} & \textcolor{red}{82.3 (0.0)} & \textcolor{red}{29.6 (-77\%)} \\ \cline{2-4} 
& 50\% & 81.6 (-0.7) & 24.6 (-81\%) \\ \cline{2-4} 
& 100\% & 77.8 (-4.5) & 19.7 (-88\%) \\ \hline \hline

\multirow{4}{*}{15\% ConvNet} & 0\% (no drop) & 78.4 (0.0) & 209.6 (0.0) \\ \cline{2-4} 
& \textcolor{red}{33\% (default)} & \textcolor{red}{78.2 (-0.2)} & \textcolor{red}{44.9 (-79\%)} \\ \cline{2-4} 
& 50\% & 74.1 (-4.1) & 37.4 (-81\%) \\ \cline{2-4} 
& 100\% & 47.1 (-31.3) & 30.0 (-86\%) \\ \hline
\end{tabular}
\caption{The trade-offs from when to drop candidates on the model accuracy and on-client communication (uplink) cost. Dynamic round number is disabled in this experiment.}
\label{tab:drop-affect}
\end{table}

\noindent \textbf{Trade-offs from drop round}
Table~\ref{tab:drop-affect} shows the trade-offs from the timing to drop the non-optimal candidates.
The results show that by dropping 33\% candidates at each round, \sys{} can reduce the uplink cost by 57\% with very little accuracy loss ($<$0.2\%).
By more aggressive early dropping, \sys{} further reduces the uplink cost, but sacrifices much more model accuracy.
In an extreme case where all non-optimal candidates are dropped immediately before the first round of model fusion (100\% drop ratio), the model accuracy degrades by 31.3\% when simplifying the ConvNet to 15\% complexity.
The reason is that with insufficient training (few rounds), the accuracy of candidates are not yet representative of the real performance of the corresponding network architectures, thus leading \sys{} to pick the wrong candidate.
The impacts from such misleading accumulate as more iterations go on.

% !TeX root = main.tex

\begin{table}[t]
\footnotesize
\begin{tabular}{|c|L{1.4cm}|L{1.5cm}|L{2.2cm}|}
	\hline
	\multicolumn{1}{|l|}{\textbf{Model}} & \textbf{Group number} & \textbf{Top-1 Accuracy (\%)} & \textbf{Avg uplink cost per client (MBs)} \\ \hline \hline
	
	\multirow{4}{*}{75\% MobileNet} & \textcolor{red}{14 (default)} & \textcolor{red}{68.8 (0.0)} & \textcolor{red}{70.3 (0.0)} \\ \cline{2-4} 
	& 7 & 68.9 (+0.1) & 140.7 (+100\%) \\ \cline{2-4} 
	& 28 & 68.6 (-0.2) & 35.2 (-50\%) \\ \cline{2-4} 
	& 100 & 68.5 (-0.3) & 9.8 (-86\%) \\ \hline \hline
	
	\multirow{4}{*}{50\% MobileNet} & \textcolor{red}{14 (default)} & \textcolor{red}{67.4 (0.0)} & \textcolor{red}{218.1 (0.0)} \\ \cline{2-4} 
	& 7 & 67.6 (+0.4) & 436.2 (+100\%) \\ \cline{2-4} 
	& 28 & 66.3 (-1.1) & 109.0 (-50\%) \\ \cline{2-4} 
	& 100 & 66.0 (-1.4) & 30.5 (-86\%) \\ \hline
\end{tabular}
\caption{The trade-offs from group number on the model accuracy and on-client communication (uplink) cost. All optimizations are enabled in this experiment.}
\label{tab:group-affect}
\end{table}

\noindent \textbf{Trade-offs from group number}
In essence, the group number determines how many clients and data are involved in training each model candidate.
As shown in Table~\ref{tab:group-affect}, with a smaller group number (7) on ImageNet, \sys{}'s accuracy doesn't improve much (up to 0.4\%) compared to our default setting (14), but incurs much more client cost (e.g., 2$\times$ more uplink network).
It confirms our observation as discussed in Section~\ref{sec:back:opt} that training and testing each model candidate only require partial clients and data to involve.
With a relatively larger group number 28, the accuracy drops by 1.1\% when adapted to 50\% complexity, but the uplink cost is also reduced by 50\%.
An even larger group number (100) helps reduce the cost by 86\% but the accuracy degradation increases up to 1.4\%.
In a word, the group number provides rich trade-offs between the generated model accuracy and on-client cost.
But note that when the group number is larger than the candidate number, further increasing it doesn't reduce the end-to-end architecture search time because of the dependency between sequential iterations.
Due to the limitation of current federated learning platforms, we currently don't evaluate this neural architecture search time and leave it as future work.

\subsection{Ablation Studies}\label{sec:eval:abla}

% !TeX root = main.tex

\begin{figure}[t]
	\centering
	\includegraphics[width=0.4\textwidth]{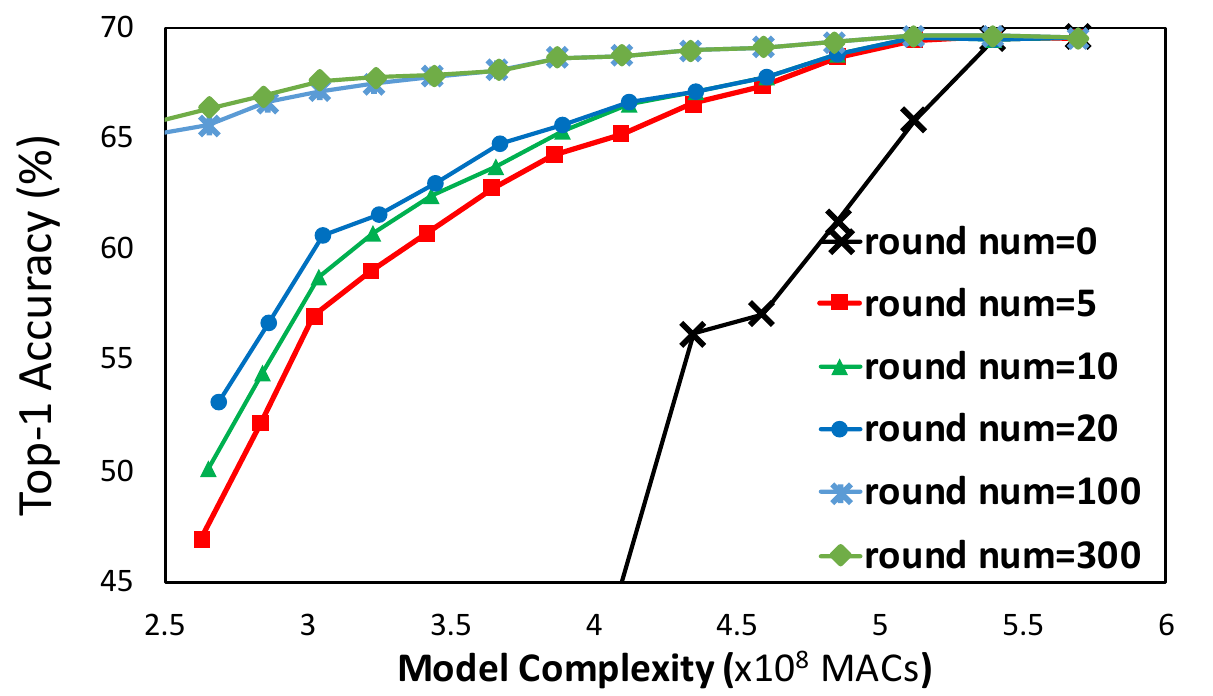}
	%\vspace*{-.6cm}
	\caption{The impacts of short-term fine-tune (round number) on model accuracy (without long-term fine-tune). Other settings are the same as Figure~\ref{fig:perf}.
				Model: MobileNet.}
	\label{fig:round}
\end{figure}
\noindent \textbf{Impact of short-term fine-tuning}
Figure~\ref{fig:round} shows the model accuracy with different round numbers (without long-term fine-tuning).
In an extreme case with zero round number, i.e., all candidates except the optimal one are dropped without model fusion, the accuracy rapidly drops to almost random guess.
In this case, the algorithm picks the best candidate solely based on noise thus gives poor performance, and the long-term fine-tune cannot save the accuracy because the model architecture is inferior.
With a reasonably smaller round number (e.g., 5 and 10), though the model accuracy can be largely preserved but still lower than the default setting.
It demonstrates that though a small round number is often enough to pick the optimal candidate at current iteration (motivation for early dropping optimization), but still we need more rounds to re-train the picked model before entering into the next round.
Otherwise the pruning direction at later iterations will be misled.

% !TeX root = main.tex

\begin{figure}[t]
	\centering
	\includegraphics[width=0.4\textwidth]{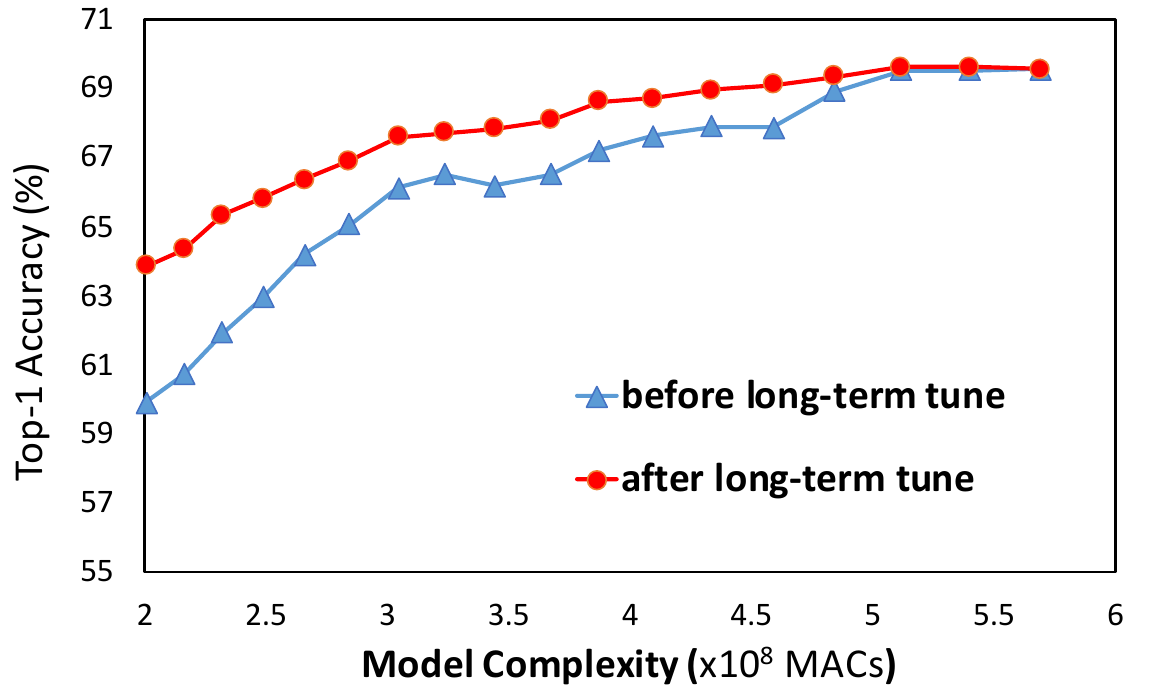}
	%\vspace*{-.6cm}
	\caption{Long-term fine-tune can substantially increase the generated model accuracy. Model: MobileNet.}
	\label{fig:longtune}
\end{figure}
\noindent \textbf{Impact of long-term fine-tuning}
Figure~\ref{fig:longtune} illustrates the importance of performing the long-term fine-tuning using federated learning after global models have been generated.
It shows that the short-term fine-tuning can preserve the accuracy well at the beginning, but the accuracy still drops faster as iterations go on due to the accumulation of insufficient training.
The long-term fine-tuning can increase the accuracy by up to another 20\% at later stages.
Though at later iterations the raw accuracy drops faster, \sys{} is still able to pick the good candidate, thus maintains close performance compared to NetAdapt as shown above.
Nevertheless, it shows that the training under the default setting has the potential to be further improved by adding more rounds.
%This is because the short-term fine-tuning has a short training time, the training is terminated far before convergence.
\section{conclusion}

In this work, we have presented a novel framework, \sys, which can automatically generate neural architectures with training data decentralized with a large number of clients.
To deal with the heavy cost of on-client computation and communication,
\sys{} identifies the key opportunity as insufficient candidate tuning by looking into the NAS intrinsic characteristics, and incorporates three key optimizations: parallel model tuning, dynamic training, and candidates early dropping.
Tested on both iid and non-iid datasets, \sys{} is able to generate neural networks with similar accuracy compared to training on centralized data, with tolerable computational and communication cost on clients.

	{\normalsize
	\bibliographystyle{acm}
	\interlinepenalty=10000 % avoid urls broken into two pages
	\bibliography{mwx.bib}
	
	\nocite{liu2007towards}

\end{document}